\documentclass{article}

\usepackage{arxiv}

\usepackage[utf8]{inputenc} 
\usepackage[T1]{fontenc}    
\usepackage{hyperref}       
\usepackage{url}            
\usepackage{booktabs}       
\usepackage{amsfonts}       
\usepackage{nicefrac}       
\usepackage{microtype}      
\usepackage{lipsum}
\usepackage{graphicx}
\usepackage[round]{natbib}
\usepackage{doi}
\usepackage{listings}
\usepackage{algpseudocode}
\usepackage{algorithm}
\usepackage{color,soul}
\usepackage{float}
\usepackage{booktabs}
\usepackage{caption}  

\algrenewcommand\algorithmicrequire{\textbf{Input:}}

\hypersetup{
    colorlinks=true,    
    linkcolor=black,    
    citecolor=black,    
    urlcolor=blue,      
}

\title{Technical Report on the Pangram AI-Generated Text Classifier}

\author{{Bradley Emi}\thanks{Equal contribution. Correspondence to \texttt{info@pangramlabs.com}} \\
	Pangram Labs\\
	Brooklyn, New York\\
	\texttt{bradley@pangramlabs.com} \\
	\And
	{Max Spero*}\\
	Pangram Labs\\
	Brooklyn, New York\\
	\texttt{max@pangramlabs.com} \\
}

\renewcommand{\headeright}{}
\renewcommand{\undertitle}{}
\renewcommand{\shorttitle}{Pangram Text Technical Report}

\hypersetup{
pdftitle={Pangram Text},
pdfsubject={q-bio.NC, q-bio.QM},
pdfauthor={Bradley Emi, Max Spero},
pdfkeywords={AI Text Detection, Large Language Models, Active Learning},
}

\begin{document}
\maketitle

\begin{abstract}
We present Pangram Text, a transformer-based neural network trained to distinguish text written by large language models from text written by humans. Pangram Text outperforms zero-shot methods such as DetectGPT as well as leading commercial AI detection tools with over 38 times lower error rates on a comprehensive benchmark comprised of 10 text domains (student writing, creative writing, scientific writing, books, encyclopedias, news, email, scientific papers, short-form Q\&A) and 8 open- and closed-source large language models. We propose a training algorithm, hard negative mining with synthetic mirrors, that enables our classifier to achieve orders of magnitude lower false positive rates on high-data domains such as reviews. Finally, we show that Pangram Text is not biased against nonnative English speakers and generalizes to domains and models unseen during training. 
\end{abstract}


\section{Introduction}

AI-generated text detection has become an increasingly important problem to solve. Large language models (LLMs) such as GPT-3, \citep{brown2020language} GPT-3.5, and ChatGPT series \citep{openai2023chatgpt}, Gemini \citep{geminiteam2023gemini}, and several open source models are now capable of producing fluent text that even trained experts are not able to distinguish perfectly \citep{CASAL2023100068}. As a result, several research and commercial solutions have recently been developed, notably TurnItIn and GPTZero \citep{tian2023gptzero}, among numerous others. 

However, the shortcomings of existing AI detection methods have been well-documented \citep{Weber_Wulff_2023}. Commercial AI detectors such as TurnItIn have demonstrated inadequate accuracy for the context of academic plagarism - primarily due to high false positive rates \citep{Vanderbilt2023}. Additionally, zero-shot AI detection methods have been shown to be biased against nonnative English writing \citep{liang2023gpt}. For many applications, such as academic integrity enforcement, reducing false positive rates is a key barrier to the adoption of AI detection tools.

Our technical approach is motivated by the need for robust detection methods with extremely low false positive rates that can be used in production settings. Perplexity based methods - such as DetectGPT, Sniffer and DetectLLM - fail on human-written documents that happen to be in the training set of LLMs, such as the Declaration of Independence. On the other hand, deep learning based classifiers are not robust to out-of-domain examples: as shown by DetectGPT and many others, these detectors perform poorly when run on text that differs significantly from their training distributions.

In this technical report, we first present a high level overview of our results as compared to two commonly used commercial detectors, GPTZero and Originality AI, as well as one of the most popular academic methods, DetectGPT \citep{mitchell2023detectgpt}. In Section 3, we explain our methodology, as well as the motivation for our technical approach: the saturation of scaling laws. We believe that the only way to train a production-level AI classifier is by training it on a large-scale dataset comparable in size to the datasets that are used to train modern LLMs. However, we show that a naive approach, simply training the classifier on a multi-million example dataset out of the box, will not work. Not only is this method extremely cost ineffective, so much so that to actually run the experiment would be impossible, but we demonstrate that at a certain critical threshold, adding additional randomly sampled examples does not improve accuracy. From our scaling law experiments, we can infer that the naive baseline would cap out in performance as well.

The solution to the scaling law saturation issues is to resolve poorly conditioned optimization (e.g. gradient collapse) via a learning curriculum based on active learning and hard negative mining. We detail a novel algorithm for training AI-text generation models that scales to web data by efficiently selecting and generating novel AI examples based on offline inference results from a partially trained classifier. We show that this approach yields superior results and enables generalization patterns that were previously not thought to be possible with a deep learning based AI-generated text classifier.

\section{Algorithm}

\begin{figure}[th]
\centering
\noindent\includegraphics[width=\textwidth]{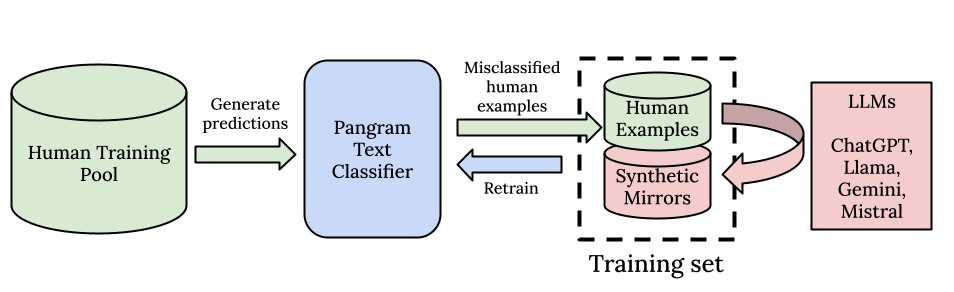}
\caption{Training process for the Pangram AI-generated text classifier. An initial classifier predicts on a large training pool of human examples, identifying false positives which are then added to the training set and mirrored by LLMs.}
\label{fig:hero}
\end{figure}
\subsection{Training Algorithm}

Our model is a slightly modified transformer-style architecture \citep{vaswani2017attention}. The classifier is trained on a mixture of human examples and synthetic examples generated by LLMs to closely match the content of the human examples, using a method called mirror prompting that we detail in Section \ref{sec:mirror_prompts}.

We consider our dataset and training method to be the primary reason for our improvement in performance over other state-of-the-art methods.

It would be exorbitantly cost-prohibitive to generate tens of millions of synthetic mirrors to match the human examples in our training pool. Additionally, we find that training a neural network with a standard loss function reaches convergence before the first epoch concludes. We hypothesize that the reason for this early convergence is that the vast majority of examples are trivially easy for the network to classify: given only a few examples, the network is able to distinguish most AI examples from most human examples. As a result, the network converges to a "spiky" loss function-- where most batches of input have close to zero loss and therefore zero gradient, drastically reducing the efficiency per unit compute. 

To solve this early convergence issue, we treat the problem as a coreset selection problem, where the optimization objective is to search for difficult examples that resuscitate the gradient signal to the network. After initial training, we use hard negative mining to ensure that we add only increasingly difficult examples to the training set to classify. Our approach also shares commonalities with curriculum learning, which rather than uniformly sampling examples from the training set, suggests training the network with a time-dependent distribution that provides better conditioning for the loss function and therefore more well-behaved optimization. A schematic of our method is presented in Figure \ref{fig:hero}. We describe the dataset and training method in more detail in Section \ref{sec:headings}.

\section{Results}

\subsection{Overview}

\begin{figure}[t]
\centering
\makebox[\textwidth]{\includegraphics[width=1.2\textwidth]{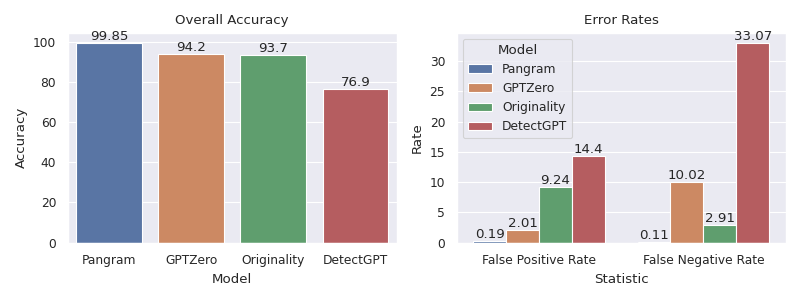}}

\caption{Overall results. (Left): Accuracy by detection method. (Right): False positive and false negative rates by detection method. Pangram has significantly higher accuracy than the next best methods, demonstrating state-of-the-art performance.}
\label{fig:overall}
\end{figure}

We present a comprehensive benchmark on 1,976 documents against commercial AI detectors GPTZero and Originality.ai, as well as the zero-shot academic method DetectGPT. The size of the dataset was chosen to balance coverage of a wide variety of domains and appropriate sample size for evaluation, with cost-effectiveness and reproducibility. The benchmark size is such that, at the time of publication, it approaches the quantity of text one can scan with a monthly subscription to GPTZero or Originality. This benchmark is comprised of documents from eight LLMs - five commercial and three open-source - as well as ten different text domains. 
\begin{table}[ht]
\centering
\begin{tabular}{ll}
\toprule
\textbf{Models} & \textbf{Domains} \\ \midrule
 & Blog Posts \\
GPT 3.5 Turbo 0301 & Scientific Writing\\
GPT 3.5 Turbo 1106 & News\\
GPT 4 0613 & Books\\
GPT 4 Turbo 1106 & Student Writing\\ 
Google Gemini Pro & Email\\
Mistral 7B Instruct & Creative Writing\\
Mixtral 8x7B Instruct & Reviews\\
LLaMA 2 70B Chat & Q\&A\\
& Wikipedia \\ \midrule
\end{tabular}
\caption{Models and text domains included in the benchmark. Each model is used to generate synthetic examples evenly across all domains.}
\label{tab:models_and_domains}
\end{table}

We chose several versions of ChatGPT because it is the dominant LLM in terms of popularity and market share \citep{HotelNewsResource2023}. While it is difficult to estimate the exact market share, based on social media analytics, ChatGPT represents 89\% of the LLM market, while Bard is a distant second at 5\%, with no other LLM surpassing 1\%. Bard has been retired, but it has been replaced with Gemini-- thus, we include Gemini Pro in our benchmark as well. We also include popular open-source models LLaMA 2 \citep{touvron2023llama} and two models from the Mistral team \citep{jiang2023mistral} \citep{jiang2024mixtral} due to their popularity in the LLM open-source community. \\

Due to its significance in AI research, we also intend to include Claude by Anthropic \citep{Anthropic2023} in a future version of this benchmark. However, we omitted it from this benchmark due to several observed issues getting Claude to respond correctly to our prompts and a high number of rejected requests. We felt that this would not be an accurate representation of model performance on Claude outputs due to the high amount of manual postprocessing that needs to be done before Claude outputs are clean enough for most practical applications. \\

For DetectGPT, we evaluate it in the most favorable black box setting for ChatGPT, using the best performing model from the large-scale study of cross detection \citep{mireshghallah2023smaller}, with 100 perturbations from T5-3B and using OPT-125M as the cross-detector.

\subsection{Overall Performance}

Pangram's text classifier is the only model that achieves production-ready levels of accuracy, false positive rate, and false negative rate. Our model is the most accurate at 99\%, compared to commercial competitors which do not even clear 95\%. Our false positive rate is better than the second best model, GPTZero, by a factor of 3, which achieving 7 times better negative error rate.

Notably, GPTZero's false negative rate is 10.02\%-- it is extremely biased towards predicting false negatives rather than false positives. This is important for educators, but seriously hampers the reliability of detection, compromising its ability to confidently authenticate that human text is actually human-written. Originality has the opposite issue-- its false positive rate is 9.24\%, which is simply too high to be practical. Human text comprises a vast majority of documents in most applications, with AI text still being a relatively novel occurrence, so with such a high false positive rate, most of the AI predicted documents will actually be human.

As DetectGPT performance is extremely poor compared to the commercial detectors on our benchmark, we omit results from DetectGPT for the remainder of the methods section.

\begin{figure}[t]
\makebox[\textwidth]{\includegraphics[width=1.2\textwidth]{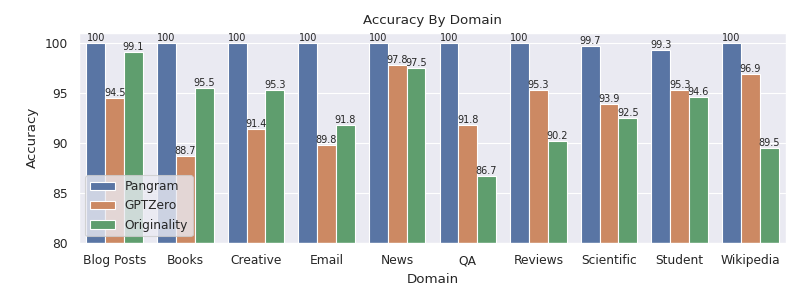}}
\centering
\caption{Accuracy by Domain. Pangram outperforms GPTZero and Originality on all 10 domains tested, demonstrating robustness to a wide variety of writing styles and formats.}
\label{fig:domain_acc}
\end{figure}

\subsection{Performance by Domain}

We compare the performance of Pangram against GPTZero and Originality on 10 domains of text: blog posts, books, creative writing, email, news, Q\&A, reviews, scientific writing, student writing, and Wikipedia. Creative writing is taken mostly from the Reddit WritingPrompts subreddit \citep{fan2018hierarchical} as well as the Ghostbusters subset \citep{verma2023ghostbuster}. Books are taken from Project Gutenberg. Email is taken from the Enron email dataset \citep{enron_email_dataset}. QA is taken from a mix of sources, including Reddit ELI5, AskHistorians, AskScience \citep{fan2019eli5}, as well as the open source MedQUAD \citep{BenAbacha-BMC-2019} and FIQA datasets \citep{fiqa2018}. Student writing is taken from a mix of sources, including the PERSUADE corpus \citep{Persuade2023}, the Liang benchmark \citep{liang2023gpt}, and college admissions essays collected from various sources around the web. Reviews are taken from Amazon, Google Maps, and Yelp directly. In all cases, human data is verified to have been written prior to 2021 to avoid contamination from AI writing. 

\begin{figure}[t]
\makebox[\textwidth]{\includegraphics[width=1.2\textwidth]{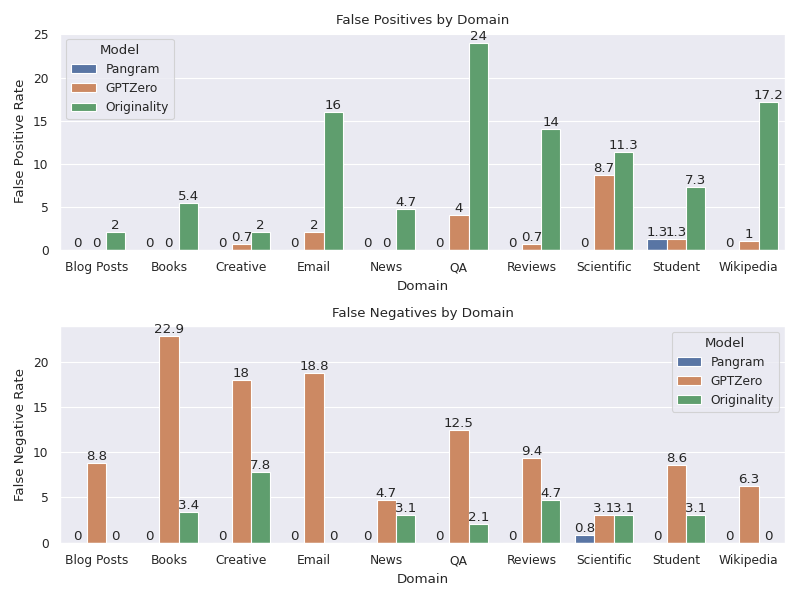}}
\centering
\caption{False Positive and False Negative Rates by Domain. Other models show bias towards over- or under-predicting AI labels. Pangram is the only model that achieves both low FPR and FNR.}
\label{fig:domain_fpr}
\end{figure}

\subsection{Performance by LLM}

We also evaluate the performance of each classifier by its ability to detect AI documents from the different LLMs in the benchmark survey. A detector with a high false positive rate could game the accuracy metric simply by lowering the threshold, capturing lots of AI text at the expense of also classifying a lot of human text as AI as well. To normalize for the threshold and avoid each classifier being able to trade off false positives for false negatives, we first set a threshold such that each model has a fixed false positive rate of 1 percent. Then, given this threshold, we evaluate the recall: the fraction of AI documents able to be classified at that threshold. Using this metric, each classifier can be evaluated independently of the chosen threshold.

GPTZero does not use a threshold, rather it uses a ternary classification scheme between "HUMAN", "AI" and "MIXED". Unfortunately it does not expose the relative probabilities of each category in the exported output, so we use the hard prediction from GPTZero instead. Since GPTZero has a false positive rate of 2 percent, these recall numbers are an optimistic upper bound on GPTZero's actual performance.

\begin{figure}[ht]
\makebox[\textwidth]{\includegraphics[width=1.2\textwidth]{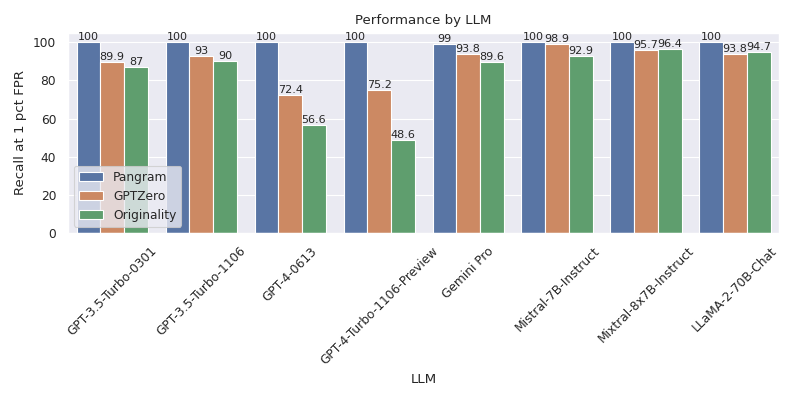}}
\centering
\caption{Recall at 1\% FPR by LLM that generated the AI text. Pangram's performance remains strong on the most capable model, GPT-4, while the other models experience a severe degradation in performance.}
\label{fig:llm}
\end{figure}

We observe that Pangram is the only model that can achieve greater than 97\% recall on all the LLMs tested. Interestingly, GPTZero performs best on the open-source models. This is likely due to the fact that perplexity and burstiness features can be used from the same model that generated the output-- as exact computation of perplexity can only be done in the white-box setting with an open source model. 

GPTZero and Originality completely fail when asked to detect AI content produced by GPT-4, both the latest full version of GPT-4 as well as GPT-4-Turbo. According to \citep{chakraborty2023counter}, the GPT-4 family of models is the \emph{most} difficult family to classify based on probability-based features, showing the limitations of the probability-based approaches to detect increasingly capable LLM output. This is the reason why Pangram uses standard deep learning techniques -- so that the model can learn the underlying patterns of speech and voice created by advanced LLMs such as GPT-4, rather than relying on probability estimates from less capable models. Additionally, the recent release of GPT-4 may mean that it was not incorporated into the training sets of the other models.

We estimate that we are the only AI detection model that will continue to hold performance even as LLMs become more capable.

\subsection{Performance on Nonnative English (ESL)}

\begin{figure}[ht]
\includegraphics[scale=0.8]{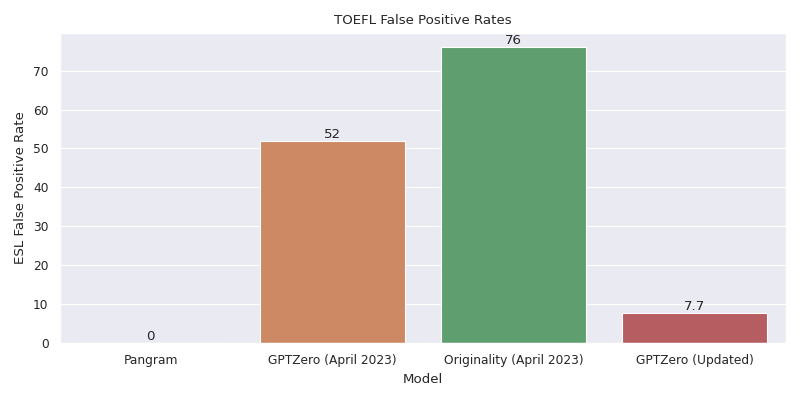}
\centering
\caption{False positive rates on the TOEFL benchmark. Pangram is the only model to achieve zero false positive rate on this English as a Foreign Language benchmark.}
\label{fig:toefl}
\end{figure}

Perhaps due to their simplistic language, researchers \citep{liang2023gpt} found that commercial LLM detectors are consistently biased against nonnative speakers (English as a Second Language, or ESL). To test this, the researchers used a benchmark of 91 essays from TOEFL (Test of English as a Foreign Language) to test several detectors. 

We hold out the 91 TOEFL essays from our training set and evaluate Pangram on the benchmark. Due to the composition of our training set, we report a false positive rate of 0\% on the TOEFL benchmark. We report results in Figure \ref{fig:toefl}.

In October 2023, GPTZero published a response to this benchmark on their blog, titled "ESL Bias in AI Detection is an Outdated Narrative" \citep{GPTZeroESL} along with an updated model. However, the new model still achieves a false positive rate of 7.7\%, or 1.1\% if "Possible AI Content" is generously labeled a negative.

We further evaluate Pangram Text on two additional ESL datasets.

We evaluate Pangram on 3,907 essays from the ELLIPSE dataset \cite{feedback-prize-english-language-learning}, which contains ESL essays from 8th-12th graders, and find zero false positives.

We also evaluate Pangram on the recently released ICNALE dataset \cite{ishikawa2023icnale}, which contains 5,600 essays from undergraduates at Asian universities learning English. We report a false positive rate of 0.09\% on this dataset.

Based on these findings, we conclude that Pangram is not biased against text written by non-native English speakers.

\subsection{Performance on Out-of-Distribution Examples}

Deep learning based detectors are often criticized for being unable to generalize outside of the text distributions that they are trained on. DetectGPT \citep{mitchell2023detectgpt} uses WMT and PubMedQA as example domains in which the RoBERTa based OpenAI text classifier fails to correctly distinguish text. While PubMedQA and WMT-like text are included in the domains (QA and general free-form text) of our classifier, one domain that is very difficult to source open data is email, due to the sensitive nature of the domain and the large amounts of personally identifiable information (PII) contained within it. 

To show that our model is able to generalize outside of its training domain, we hold out all email from our training set and evaluate our model on the entire Enron email dataset, which was released publicly as a dataset for researchers following the extrication of the emails of all Enron executives in the legal proceedings in the wake of the company's collapse.

Our model with email held out achieves a false positive rate of 0.8\% on the Enron email dataset after hard negative mining, compared to our competitors (who may or may not have email in their training sets) which demonstrate a FPR of at least 2\%. After generating AI examples based on the Enron emails, we find that our false negative rate is around 2\%. We show an overall accuracy of 98\% compared to GPTZero and Originality which perform at 89\% and 91\% respectively. 

After retraining on all Enron email, except for a test set of 1000 held-out Enron email documents, we are able to achieve zero errors on the test set.

\subsection{Performance on LLMs unseen during training}

We are aware that we won't be able to train on the outputs of every model or even model family. However, our classifier must perform well on new and out-of-distribution open source models if we expect to serve as an effective safeguard for the internet. We evaluate the Pangram classifier on evaluation sets of 1300 examples from each of several models available in the Together AI API. The results in Table \ref{tab:open_source_models} show that our model effectively generalizes to open source models, with greater than 99.6\% recall on all models tested.

\begin{table}[ht]
\centering
\begin{tabular}{lc}
\toprule
\textbf{LLM} & \textbf{Recall at 1\% FPR} \\
\midrule
OpenChat 3.5 & 100.00\% \\
Qwen1.5-72B-Chat & 99.93\% \\
DeepSeek-Coder-33B-Instruct & 99.75\% \\
Yi-34B-Chat & 99.68\% \\
Vicuna-13B-v1.5 & 99.85\% \\
MythoMax-L2-13b & 99.61\% \\
SOLAR-10.7B-Instruct-v1.0 & 99.61\% \\
\bottomrule
\end{tabular}
\caption{Pangram performance on open source models unseen during training. Our model shows robustness to new models and model families trained on open source data.}
\label{tab:open_source_models}
\end{table}

\subsection{July 2024 Update: Performance on Recently Released LLMs}

We evaluated Pangram Text on the major recently released commercial large language models, as of July 2024, using a benchmark of 25,000 documents generated in the same way as the original benchmark. We observe that we are able to continue achieve excellent performance on these newly released models.

\begin{table}[ht]
\centering
\begin{tabular}{lc}
\toprule
\textbf{LLM} & \textbf{Recall at 1\% FPR} \\
\midrule
GPT 4o & 100.0\% \\
Claude 3 (various sizes) & 99.76\% \\
LLaMA 3 (various sizes) & 99.97\% \\
\bottomrule
\end{tabular}
\caption{Pangram performance on recently released models from July 2024.}
\label{tab:new_models}
\end{table}

\subsection{July 2024 Update: Performance on non-English Languages}

In July 2024 we updated Pangram text to distinguish multilingual human and AI-generated text. To benchmark our performance, we use the Amazon Multilingual Reviews dataset \citep{keung2020multilingual}, the XLSum dataset \citep{hasan2021xl}, and the multilingual Wikipedia datasets, and generate synthetic mirrors for each.

\begin{table}[ht]
\centering
\begin{tabular}{lccc}
\toprule
\textbf{Language} & \textbf{Amazon Reviews Accuracy} & \textbf{Wikipedia Accuracy} & \textbf{XLSum (BBC News) Accuracy} \\
\midrule
Spanish & 99.59\% & 99.75\% & 99.75\% \\
French & 98.84\% & 99.33\% & 98.50\% \\
Italian & N/A & 99.82\% & N/A \\
German & 99.44\% & 99.95\% & N/A \\
Portuguese & N/A & 99.83\% & 99.70\% \\
Russian & N/A & 98.34\% & 99.35\% \\
Chinese & 99.70\% & 99.54\% & 98.10\% \\
\hline
\end{tabular}
\caption{Pangram multilingual performance.}
\label{tab:multilingual}
\end{table}

\subsection{Additional Benchmark Information}

\subsubsection{Dataset Sources}

Our benchmark dataset comprises examples both from open-source text corpora as well as hand-picked examples from the internet. The importance of using hand-picked examples is that we do not know the training sets of other commercial detectors, but it is likely that they are trained on common open-source datasets available on HuggingFace and other common repositories. To accurately measure real-world performance and mitigate the risk of training set leakage, we collect text from many websites. For example, for the news domain, we use a fraction of the Reuters, BBC, and XSum datasets, but we also use a fraction directly copied and pasted from individual international, national, and local news websites. We publicly release our dataset \footnote{\texttt{https://checkforai-public.s3.amazonaws.com/benchmark.csv}} and can provide an exhaustive list of sources upon request.

\subsubsection{AI-Generated Examples}

We use standard prompts to create matching AI examples for each human example. For example, "Write a title for the following essay" followed by "Write an essay with the following title." We sometimes ask the LLM to use the first sentence of the document as the first sentence of the response, similar to \citep{mitchell2023detectgpt}. 

We attempt to remove standard LLM boilerplate headers, such as "Sure! Here is an essay in response to your prompt:", by using the following heuristic. If the first paragraph starts with any of the following phrases: \texttt{Sure}, \texttt{Here is a}, \texttt{Title:}, \texttt{Abstract:}, \texttt{I have:}, or \texttt{I'm happy to help}, then we remove it. We reject any examples under 50 words long as these are typically implicitly rejected requests.

Finally, to avoid having any of the models using formatting rather than content to make their predictions, we remove all non-Unicode characters using \texttt{unidecode}, normalize all whitespace and quotations, and remove emojis. The result is a very clean benchmark of AI-generated examples that accurately reflect detection of AI-generated text based on content.





\section{Method}

\label{sec:headings}

\subsection{Datasets}

To train our model, we begin by considering a total pool of approximately 28 million confirmed human-written documents, described broadly in Table \ref{tab:domain_examples}. The datasets are open source, freely available on the Internet, and importantly, are licensed for commercial use. We only consider datasets from 2021 and earlier to minimize the likelihood that any AI-generated text ends up mislabeled in our training set.

We exclude 4 million examples from our training pool as a holdout set to evaluate false positive rates following calibration on the above benchmark.

\begin{table}[ht]
\centering
\begin{tabular}{lc}
\toprule
\textbf{Domain} & \textbf{Number of Examples} \\
\midrule
Creative Writing       & 300,000 \\
Business and Product Reviews & 15,000,000 \\
Books               & 7,000,000 \\
Scientific Papers       & 3,000,000 \\
Wikipedia         & 1,000,000 \\
News Articles & 500,000 \\
Q\&A & 1,000,000 \\
Email & 16,000 \\
Student Writing & 23,000 \\
English as a Second Language (ESL) & 165,000 \\
\midrule
\end{tabular}
\caption{Human-Authored Examples Available by Domain}
\label{tab:domain_examples}
\end{table}

\subsection{Mirror Prompts}
\label{sec:mirror_prompts}

A naive implementation of an AI text detection classifier could put $N$ human documents into a training set and prompt an LLM for $N$ AI-generated documents. A pitfall with this approach is that there are a number of features to overfit to - the topic chosen by the LLM, the length of the document, the default level of sophistication.

In order to prevent overfitting to any of these content-specific features, we design the AI side of the dataset to closely resemble the human side in style, tone, and semantic content. For each human example, we generate an AI-generated example that matches the original document on as many axes as possible, to ensure that our model learns to classify documents solely based on specific characteristics of LLM writing.

We define the term "mirror prompt" to be a prompt based on the original example that is used to generated a "synthetic mirror" or "mirror example." The goal of each mirror prompt is to generate an example that matches the topic and length of the original document.

If the original document is "<original review>", then a mirror prompt may look like this:

\begin{lstlisting}
[Prompt] Write a <original review star rating> star review for <original review business name>. Make the review around <original review length> words long. 
\end{lstlisting}

Another example may be for a student essay. We sometimes use double prompts, such as the following:

\begin{lstlisting}
[Prompt] What is a good title for this essay? <original essay> Only give the title in your response.

[Assistant] <Title>

[Prompt] Write an essay with the following title: <Title>. Make the essay around <original essay length> words long.
\end{lstlisting}

Another goal of mirror prompting is to increase robustness. A simple way to do this is to prompt the LLM to complete a document, starting with an excerpt from the document. This strongly steers the LLM to write in a style matching the original document, which makes the task harder and improves our learning signal.

An example mirror prompt in this style is:
\begin{lstlisting}
[Prompt] Write a Wikipedia article about <topic>. Start with these sentences: <excerpt>. Make the article about <original article length> long. 
\end{lstlisting}

Mirror prompts must be hand-tuned for each domain to remove obvious AI tells. For example, when asking for an essay, LLMs often include as the first line a title or "Sure, here is an essay about <topic>: ". Removing these artifacts is vital, otherwise the classifier will simply learn to look only for these tells. This is often done by appending an instruction to the mirror prompt:
\begin{lstlisting}
Do not include a title, word count, or any information besides that of the actual essay.
\end{lstlisting}

In addition to adding this instruction to the prompt side, we also have postprocessing filters that look for the most obvious tells and discard common patterns when found.

\subsection{Scaling Laws}

\begin{figure}[ht]
\includegraphics[scale=0.65]{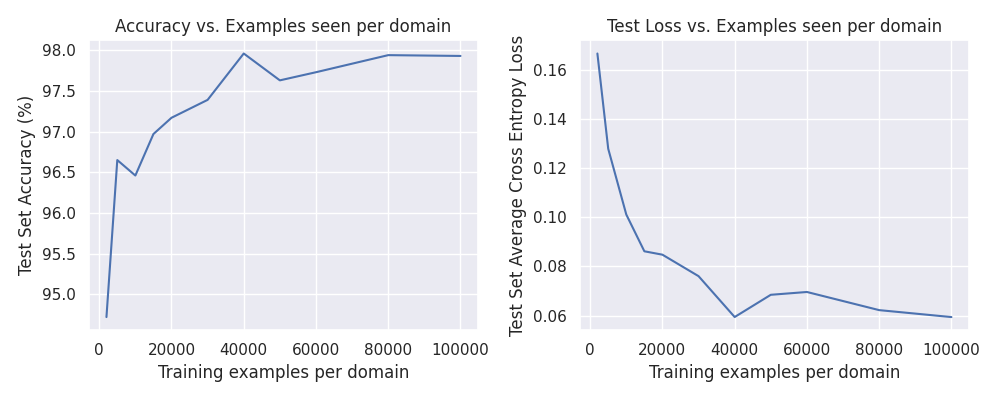}
\centering
\caption{Scaling laws: loss and accuracy against number of training examples seen. Performance begins to saturate around 40,000 examples per domain.}
\label{fig:scaling}
\end{figure}

As an initial experiment, we train several classifiers on a series of increasingly large, randomly sampled training sets. Checkpoints are selected based on validation loss and then test loss and test accuracy for each dataset size are computed using that checkpoint. We show in Figure \ref{fig:scaling} that performance reaches saturation around 40,000 examples per domain (including both real examples and mirrors), which is equivalent to around 400,000 examples total. Even at 100,000 examples per domain, which is 1 million total examples, after saturation has been reached, we see that the model cannot converge to an accuracy above 98\%.  

\subsection{Hard Negative Mining with Synthetic Mirrors}
\label{fig:hard_neg}
In order to overcome the increasing inefficiency of training a neural network on an extremely large scale dataset of 28 million examples, we developed a hard negative mining algorithm, described below in Algorithm 1. We alternate training a series of neural networks to convergence, with rounds of hard negative mining and synthetic mirror data generation in between in order to surface difficult examples that the model can more efficiently learn from.

\begin{algorithm}[ht]
\caption{Hard Negative Mining with Synthetic Mirrors}
\label{alg:hard_negative_mining}
\begin{algorithmic} 
\Require Large training pool of human examples $\mathbb{P}$, initial training set size $n$, round size $m$.
\State Initialize training set $T_0 \gets n $ examples randomly sampled from $\mathbb{P}$.
\State For each example in $T_0$, generate mirrored synthetic examples and add them to $T_0$.
\State Train initial model $M_0$ on $T_0$ to saturation. 
\State $i \gets 0$
\While{$M_i$ is better than $M_{i-1}$}
    \State Generate predictions for each example in $\mathbb{P}$ using current model $M_i$.
    \State Randomly sample $m$ false positives $F_i$ from predictions.
    \State For each example in $F_i$, generate mirrored synthetic examples $S_i$.
    \State Create a new training set $T_{i+1} \gets T_i \cup F_i \cup S_i$.
    \State Train model $M_{i+1}$ on the updated training set $T_{i+1}$.
    \State $i \gets i + 1$
\EndWhile
\end{algorithmic}
\end{algorithm}

We follow Algorithm \ref{alg:hard_negative_mining}, hard negative mining with synthetic mirrors, to train our model. Using our scaling laws experiment to inform our choice of initial training pool, we begin with a training set size of 40,000 examples per domain ($n=360,000$ total), the smallest number required to reach saturation. 

For each domain (except student writing - not enough examples), we select up to 10,000 false positives ($m=80,000$), generate an equal number of synthetic mirrors, add them to the training set and retrain the model.

To track our performance accurately even at very low false positive rates, we hold out approximately 4,000,000 examples from our training pool. These examples are excluded from hard negative mining rounds in order to evaluate the model at each step. The performance of the model on the holdout set implies that hard negative mining surfaces difficult examples that our model can use to generalize, and is not just memorizing hard examples until there are no more remaining.

\begin{table}[b]
\centering
\begin{tabular}{lcc}
\toprule 
\textbf{Domain} & \textbf{Base Model} & \textbf{After Hard Negative Mining} \\
\midrule 
Creative Writing & 1.51\% & 0.02\% \\
Reviews & 1.81\% & 0.02\% \\
Books & 0.85\% & 0.01\% \\
Scientific Papers & 1.54\% & 0.04\% \\
Wikipedia & 5.34\% & 0.05\% \\
News & 0.55\% & 0.001\% \\
Q\&A & 2.52\% & 0.009\% \\
ESL & 1.44\% & 0.01\% \\
\midrule 
Student Writing (No Hard Negative Mining) & 0.75\% & 0.04\% \\ 
Email & 6.60\% & 0.00\% \\
\midrule 
Total domain-weighted FPR & 2.29\% & 0.02\% \\
\midrule 
\end{tabular}
\caption{False positive rate on held out hard negative mining sets, by domain. Hard negative mining reduces false positive rates by 100x-1000x on holdout sets.}
\label{table:hard_negative_mining}
\end{table}

The results in Table \ref{table:hard_negative_mining} show that this process improves false positive rate across domains, and also results in better performance on email, a domain unseen by either model. The fact that performance on student writing does not change, is indicative that our model is limited by a lack of large-scale training data in this domain and likely has room for improvement in the future. 

Interestingly as well, after hard negative mining on other domains, we find that the performance on email (which we intentionally exclude from the training set to demonstrate out-of-domain performance) significantly improves. This suggests that the other domains have a symbiotic relationship with the email domain-- adding examples from other domains helps email as well. Therefore, we can expect that adding more domains and scaling up the size our initial training pool even further has the potential to not only cover more domains of text but also improve the model's overall generalization ability.

\section{Related Work}

\subsection{Hard Negative Mining}

Hard negative mining has long been used as a technique for object detection in computer vision, as the problem of detecting semantically meaningful objects in images is severely class imbalanced (images are mostly background and therefore most patches are too easy to classify).  \citep{lsvm-pami} \citep{shrivastava2016training}. Our work is related to this problem although we are operating in the language domain rather than the image domain, because most documents are quite easy to classify. Human documents, such as reviews, are often all very similar to each other, and AI documents often have dead giveaways (such as overusing phrases like "delve" and "it is important to note" or "as an AI language model") and so once the model learns from a few examples that a \emph{category} of text is likely to be human or AI, it does not need to see more examples from the category.

\subsection{Deep Active Learning}

Deep active learning is also a well-researched field of study and an excellent survey can be found here \citep{DBLP:journals/corr/abs-2009-00236}. Active learning is used when the learning algorithm can interactively query an information source to label new data points. A salient example of active learning being used in a detection task is in \citep{Aghdam2019} where the neural network's mispredictions of bounding boxes in an object detection task is used to query human labelers for frame-level annotations. While we do not use the learning algorithm to query LLMs to give us new labels, our work is related to active learning because we use the underlying neural network to \emph{prompt} the LLMs (the information source) for novel synthetic examples, which guides and updates our training set towards balance that promotes well-conditioned stochastic optimization. Active learning is often used to modify the "working set" of training examples that a classifier such as an SVM or a neural network is trained on. We use our active prompting scheme to augment and improve our working set of examples.

\subsection{Probability-Based Feature Classifiers}

Probability-based Feature Classifiers are based on the assumption that text generated by an LLM is generally more likely under the model's probability distribution than similar text written by a human. GLTR \citep{gehrmann2019gltr} uses entropy, probability, and probability rank to determine whether or not the statistical patterns in the text are AI-generated. The original version of GPTZero \citep{tian2023gptzero} was based on both perplexity and burstiness: the variation of perplexity across the token sequence. DetectGPT \citep{mitchell2023detectgpt} uses perturbations made by a T5 model to compare the probability of a source text to perturbed texts, finding that source texts with a high probability discrepancy against their perturbations are likely to be LLM generated. DetectGPT is a white-box method, meaning that the original LLM used to generate the text is known. However, it was recently shown that smaller language models can be used as proxy models for DetectGPT with reasonable accuracy \citep{mireshghallah2023smaller}.  DetectLLM \citep{su2023detectllm} builds upon DetectGPT by also introducing similar methods based on log-rank information. Sniffer \citep{li2023origin} uses perplexity-based features to detect AI-generated text and also which LLM produced it.

\subsection{Deep Learning Based Classifiers}

Several deep learning based classifiers have also emerged as alternatives to probability based classifiers. OpenAI's classifier \citep{openai2023classifier} is a RoBERTa based deep learning model finetuned to predict whether a document is AI-written or not based on a large training set composed of both open-source pretraining data and AI-written data generated by OpenAI's API. Due to its low rate of accuracy, it was discontinued in 2023. More recently, SeqXGPT \citep{wang2023seqxgpt} is an example of a hybrid classifier that uses both probabilistic and deep learning elements to show strong performance on sentence level AI classification. It uses extracted perplexities from a variety of language models, and then runs them through a relatively shallow neural network to predict whether sentences are ChatGPT-generated or not. Ghostbuster \citep{verma2023ghostbuster} is another recent work that trains a shallow classifier to detect AI-generated text based on combinations of unigram, trigram, and GPT-3.5 embedding features. A recent study \citep{pu2023zeroshot} has shown that deep learning based AI-generated text classifiers exhibit some generalization ability to detect text from larger models when trained on outputs of smaller models from the same family. Finally, RADAR \citep{hu2023radar} uses an adversarial learning approach to increase the robustness of a deep learning based AI detector to paraphrasing attacks. While showing significant progress in detection robustness, like DetectGPT, it is a white-box method, meaning a different detector must be used for each individual LLM, limiting its effectiveness in practice.

\subsection{Watermarking}

Watermarking is another proposed framework to mitigate harmful usage of AI-generated text. In this setting, the LLM itself leaves behind a probabilistic signal that makes it significantly easier to detect. The watermark proposed by Kirchenbauer et. al \citep{kirchenbauer2023watermark} introduces the idea of "green tokens": using some tokens at much higher frequencies than their expectation, allowing a detector with knowledge of the green token set to count the frequency of green tokens and evaluate it against random chance. While watermarking LLMs is an important research topic, with several unwatermarked LLMs already open source and in the wild, the ability to regulate and enforce watermarking regulations as of the time of publication is extremely limited, making a detection solution based on watermarking currently impractical. Furthermore, recent studies \citep{sadasivan2023aigenerated} \citep{krishna2023paraphrasing} have shown that watermarking schemes can easily be bypassed using another model that can paraphrase the output of an LLM. 

\section{Conclusion}

In this technical report we have detailed the performance of the Pangram text classifier against other commercial AI detection systems. We have shown that a simple algorithm for hard negative mining combined with a synthetic mirroring strategy is necessary to scale the performance of a deep-learning based classifier to industry-ready levels of accuracy. We show that the algorithm's novel ability to scale to web-sized datasets enables generalization to out-of-domain text, LLM-generated text where the LLM has not been observed during training, and ability to detect AI-generated text from even the current most capable model, GPT-4. 

In future work, we will continue our efforts to increase the robustness of our classifier to adversarial attacks, scale our datasets to even larger sizes, and improve the architecture and training algorithm to further improve the accuracy and generalization ability of the model.

\section{Ethics and Responsible Use}

Our classifier is available for public commercial use at \texttt{https://pangram.com}. We strongly discourage the use of our classifier as a sole arbiter of academic integrity and plagiarism checking. All AI detection tools have a nonzero false positive rate, and should be used in conjunction with other evidence to prove or disprove plagiarism. 

AI detection is not a substitute nor a reliable tool for proving the factuality or verity of textual information such as news and media. While AI is often used for disinformation, scams, or other malicious activities, human-authored text can also be used with malintent. One should always check the sources of any text in publications for truth independently.

\bibliographystyle{unsrtnat}
\bibliography{references}  






\end{document}